\documentclass[journal]{IEEEtran}

\input{xhc.sty}

\begin{document}

\title{Arbitrage of Energy Storage in Electricity Markets with Deep Reinforcement Learning}

\author{Hanchen Xu, Xiao Li, Xiangyu Zhang, and Junbo Zhang
\thanks{Hanchen Xu and Xiao Li are with the University of Illinois at Urbana-Champaign, Urbana, IL 61801, USA. (email: \{hxu45, xiaoli20\}@illinois.edu)}
\thanks{Xiangyu Zhang is with the Department of Electrical and Computer Engineering, Virginia Tech, Arlington, VA, 22203, USA. (email: zxymark@vt.edu)}
\thanks{Junbo Zhang is with School of Electric Power, South China University of Technology, Guangzhou 510641, China. (Corresponding author, email: epjbzhang@scut.edu.cn)}
}

\maketitle

\begin{abstract}
In this letter, we address the problem of controlling energy storage systems (ESSs) for arbitrage in real-time electricity markets under price uncertainty.
We first formulate this problem as a Markov decision process, and then develop a deep reinforcement learning based algorithm to learn a stochastic control policy that maps a set of available information processed by a recurrent neural network to ESSs' charging/discharging actions.
Finally, we verify the effectiveness of our algorithm using real-time electricity prices from PJM.
\end{abstract}

\begin{IEEEkeywords}
electricity markets, energy storage, arbitrage, deep reinforcement learning, recurrent neural network.
\end{IEEEkeywords}

\section{Introduction}

\IEEEPARstart{E}{nergy} storage systems (ESSs) can significantly enhance power system flexibility through the provision of multiple services in  electricity markets.
Yet, it is necessary to identify revenue sources for ESSs so as to encourage their participation in electricity markets \cite{eyer2010energy}.
Under existing market schemes, one major revenue source for ESSs is arbitrage in electricity markets \cite{krishnamurthy2018energy, wang2018energy}.
The arbitrage problem of ESSs has been studied in many existing works, such as \cite{krishnamurthy2018energy} where scenario-based stochastic optimization is applied for arbitrate between the day-ahead and real-time markets, and \cite{wang2018energy} in which the Q learning algorithm is utilized for arbitrage across different hours within the real-time market.

In this letter, we focus on the arbitrage problem of ESSs across different hours within the real-time market.
We propose a deep reinforcement learning (DRL) based algorithm to learn a stochastic control policy that maps a set of available information to ESSs' charging/discharging actions.
We first model this problem as a Markov decision process (MDP), where the state is constructed from available information, motivated by the idea developed in our earlier work in \cite{xu2019deep}.
In particular, we use an exponential moving average (EMA) filter and a recurrent neural network (RNN) to extract useful information from the sequence of electricity prices and include it in the state.
The optimal policy that solves the MDP is found using a state-of-the-art DRL algorithm---the proximal policy optimization (PPO) algorithm \cite{schulman2017proximal}.

\section{Problem Formulation} \label{sec:algo}

In this section, we develop an MDP model (see, e.g., \cite{sutton2018reinforcement} for the definition of MDPs) for the arbitrage process of an ESS.
Throughout this letter, we use a subscript~$t$ to denote the value of a variable at time instant~$t$.
Let $\tau$ denote the duration between two time instants.

\noindent 1) \textit{State Space}:
Let $E$ denote the remaining energy of the ESS, where $0 \leq \underline{E} \leq E \leq \overline{E}$.
In addition, let $p^c$ and $p^d$ denote the charing and discharging powers of the ESS, and $\overline{p}^c$ and $\overline{p}^d$ the maximum charging and discharging powers.
The charging and discharging efficiencies are denoted by $\eta^c$ and $\eta^d$, respectively.
The state transition of the ESS can be characterized as follows:
\begin{align} \label{eq:energy_trans}
	E_{t+1} = E_t + (p^c_t - p^d_t) \tau,
\end{align}
where $E_1$ is set to $\underline{E}$.
Let $\rho$ denote the electricity price, and define a function $\phi$ that extracts a hidden state $\bm{h} \in \bbR^n$ from the electricity prices as follows:
\begin{align} \label{eq:price_trans}
	\bm{h}_{t+1} = \phi(\bm{h}_t, \rho_{t+1}).
\end{align}
The hidden state $\bm{h}_t$ is expected to provide more information (such as the trend) of electricity prices in addition to $\rho_t$ itself.
The choice of $\phi$ will be detailed later in Section \ref{sec:algo}.
We next introduce the average energy cost, denoted by $c$, which only changes when the ESS charges:
\begin{align} \label{eq:cost_trans}
	c_{t+1} = \frac{c_t E_t + \rho_t p^c_t \tau / \eta^c}{E_t + p^c_t \tau},
\end{align}
where $c_1$ is set to $0$.
Note that \eqref{eq:cost_trans} does not hold when $E_{t+1} = 0$, in which case $c_{t+1}$ is set to $0$.
The state at time instant $t$ is defined as $\bm{s}_t = (E_t, c_t, \rho_t, \bm{h}_t)$ and the state space is $\calS = \{\bm{s}\} = [\underline{E}, \overline{E}] \times \bbR \times \bbR \times \bbR^n$.

\noindent 2) \textit{Action Space}:
As shown by authors in \cite{wang2018energy}, the optimal value of $p^d_t$ lies in $\{0, \min(\overline{p}^d, (E_t - \underline{E})/ \tau) \}$ and that of $p^c_t$ lies in $\{0, \min (\overline{p}^c, (\overline{E} - E_t)/ \tau )\}$; moreover, at most one of $p^d_t$ and $p^c_t$ can be nonzero.
Therefore, we define the action space as $\calA = \{ a \} = \{1, 2, 3\}$, the element in which respectively corresponds to discharging at $\min (\overline{p}^d, (E_t - \underline{E})/ \tau)$, charging at $\min (\overline{p}^c, (\overline{E} - E_t)/ \tau)$, and neither discharge nor charge.

\noindent 3) \textit{Reward}: 
The design of a reward function is crucial in MDPs.
In this problem, the reward received after taking action $a_t$ in state $s_t$, denoted by $r_t$, is defined as follows:
\begin{align} \label{eq:reward}
	r_t = 
	\left\{ \begin{array}{ll}
	(\rho_t \eta^d - c_t) p^d_t \tau -\beta p^d_t, & a_t = 1, \\
	-\beta p^c_t, & a_t = 2, \\
	0, & a_t = 3,
	\end{array} \right.
\end{align}
where $\beta > 0$ is in \$/MW, representing the per-unit wear-and-tear cost.
Except the charing/discharging cost, the ESS only incurs a profit/loss of $(\rho_t \eta^d - c_t) p^d_t \tau$ when it discharges; this reward function acknowledges the economic value of the remaining energy of the ESS.
Indeed, $\sum_{t=1}^T (\rho_t \eta^d - c_t) p^d_t \tau$ is the cumulative profit/loss incurred by the ESS by arbitrage over $T$ time instants, which we will use as a meaningful metric to evaluate the performance of the arbitrage algorithm.

\noindent 4) \textit{Policy}:
Due to the discrete nature of the action space, we adopt a categorical policy, denoted by $\pi$, as the ESS control policy.
Specifically, $\bm{s}$ is mapped to $\mu(s) \in \bbR^{|\calA|}$, where ${|\cdot|}$ indicates the cardinality of a set, via a function $\mu$ that is parameterized by $\bm{\theta}$.
Let $\mu_i(s)$ denote the $i^{\text{th}}$ entry of $\mu(s)$, then the probability of choosing action $a \in \calA$ at state $s$, denoted by $\pi(a | s)$, is the following:
\begin{align} \label{eq:policy}
	\pi(a = i | s) = \frac{e^{\mu_i(s)}}{\sum_{i=1}^{|\calA|} e^{\mu_i(s)}}.
\end{align}
The action is sampled according to \eqref{eq:policy}. 
The goal is to find $\theta$ that maximizes the expected cumulative discounted reward $\expect{\sum_{t=1}^\infty \gamma^{t-1} r_t}$, where $\gamma \in [0, 1)$ is a discount factor.
This is achieved via the PPO algorithm to be detailed next.

\section{Algorithm} \label{sec:algo}


\subsection{Hidden State Extraction}

The hidden state extractor $\phi$ is implemented via an EMA filter and an RNN\footnote{More advanced architectures of RNNs such as the long short-term memory (LSTM) can be readily used here to define the feature mapping.}, which take a sequence of electricity prices $\{\rho_1, \cdots, \rho_T\}$ as the input.
Specifically, the sequence of hidden states $\{\bm{h}_t\}$ is generated as follows:
\begin{align}
\tilde{\rho}_{t+1} &= \alpha \tilde{\rho}_t + (1 - \alpha) \rho_{t+1}, \label{eq:EMA} \\
\bm{h}_{t+1} &= \tanh(\bm{W} \bm{h}_t + \bm{w} \tilde{\rho}_{t+1} + \bm{b}), \label{eq:RNN}
\end{align}
where $\alpha \in [0, 1]$, $\tilde{\rho}_1 = \rho_1$, $\bm{h}_0$ is randomly initialized, $\tanh(\cdot)$ is applied element-wise, $\bm{W} \in \bbR^{n\times n}$ and $\bm{w} \in \bbR^n$ are unknown weights, $\bm{b} \in \bbR^n$ is an unknown bias vector.
The vector $\bm{h}_t$ is related to $\hat{\rho}_{t+1}$---an estimate of $\tilde{\rho}_{t+1}$---via 
\begin{align} \label{eq:RNN_out}
	\hat{\rho}_{t+1} = (\bm{w}^o)^\top \bm{h}_{t} + (b^o)^\top,
\end{align}
where $\bm{w}^o \in \bbR^n$ is a weight vector and $b^o \in \bbR$ is a bias.
The values of $\bm{W}, \bm{b}, \bm{w}^o, b^o$ can be optimized by minimizing $\sum_{\text{seq.}} \sum_{t=2}^{T} ( \hat{\rho}_t - \tilde{\rho}_t)^2$, where the first summation is taken over all input sequences, using backpropagation through time \cite{goodfellow2016deep}.

The EMA filter filters out high frequency components in the electricity prices, and then the RNN extracts a hidden state that is sufficient for predicting the next smoothed electricity price.
Essentially, $\phi$ aims to extract a hidden state which, together with the up-to-date electricity price, is sufficient to characterize the dynamic behavior of the electricity price sequence.

\subsection{Policy Learning}

Before introducing the PPO algorithm, we review the state value function, the action value function, and  the advantage function under policy $\pi$, defined as $V^\pi(\bm{s}_t) = \expect{\sum_{l=0}^\infty \gamma^l r_{t+l} | \bm{s}_t}$, $Q^\pi(\bm{s}_t, a_t) = \expect{\sum_{l=0}^\infty \gamma^l r_{t+l} | \bm{s}_t, a_t}$, and $A^\pi(\bm{s}_t, a_t) = Q^\pi(\bm{s}_t, a_t) - V^\pi(\bm{s}_t)$, respectively.
Intuitively, the state (action) value function indicates how good the state (state-action pair) is in the long-term, and the advantage function measures how much better the action is than average.

We write $\pi_{\bm{\theta}}$ to emphasize the fact that $\pi$ is characterized by $\bm{\theta}$.
Instead of optimizing $\bm{\theta}$ for maximizing the cumulative discounted reward, the PPO algorithm improves the value of $\bm{\theta}$ iteratively by maximizing a surrogate objective function. 
Let $\bm{\theta}_k$ denote the value of $\bm{\theta}$ at iteration $k$.
Then, the PPO algorithm improves $\bm{\theta}$ iteratively as follows:
\begin{align} \label{eq:PPO}
	\bm{\theta}_{k+1} = \argmax_{\bm{\theta}} \underset{\bm{s}, a \sim \pi_{\bm{\theta}_{k}}} \bbE[L(\bm{s}, a, \bm{\theta}_{k}, \bm{\theta})],
\end{align}
where $L(\bm{s}, a, \bm{\theta}_{k}, \bm{\theta}) = \min (\frac{\pi_{\bm{\theta}}(a | \bm{s})}{\pi_{\bm{\theta}_{k}}(a | \bm{s})} A^{\pi_{\bm{\theta}_{k}}}(\bm{s}, a), g(\epsilon, A^{\pi_{\bm{\theta}_{k}}}(\bm{s}, a)) )$, and $g(\epsilon, A)$ equals to $(1+\epsilon) A$ if $A \geq 0$ and $(1-\epsilon) A$ if $A < 0$.
If we collect $D$ state transition trajectories by running policy $\pi_{\bm{\theta}_k}$ for $T$ time instants in each trajectory, then we can approximate the expectation in \eqref{eq:PPO} by a sample average, and replace \eqref{eq:PPO} by
\begin{align} \label{eq:policy_update}
	\bm{\theta}_{k+1} = \argmax_{\bm{\theta}} \frac{1}{D T} \sum_{\text{trajectory}} \sum_{t=1}^{T} L(\bm{s}_t, a_t, \bm{\theta}_{k}, \bm{\theta}),
\end{align}
where the first summation is taken over $D$ trajectories.

To get an estimate of the advantage function that appears in the surrogate function $L$, we need to first estimate the state value function.
Let $\hat{V}_{\bm{\psi}}^\pi$ denote an estimate of $V^\pi$ that is parameterized by $\bm{\psi}$.
Let $\bm{\psi}_k$ denote the value of $\bm{\psi}$ at iteration $k$, then $\bm{\psi}_k$ can be estimated by solving
\begin{align} \label{eq:state_value_func}
	\bm{\psi}_k = \underset{\bm{\psi}}{\argmin} \frac{1}{D T} \sum_{\text{trajectory}} \sum_{t=1}^{T} \norm{\hat{V}_{\bm{\psi}}^{\pi_{\bm{\theta}_{k}}}(\bm{s}_t)-\tilde{V}^{\pi_{\bm{\theta}_{k}}}(\bm{s}_t)}^{2},
\end{align}
where $\tilde{V}^{\pi_{\bm{\theta}_{k}}}(\bm{s}_t) = \sum_{l=0}^{T - t - 1} \gamma^l r_{t + l} + \gamma^{T-t} \hat{V}^{\pi_{\bm{\theta}_{k-1}}}(\bm{s}_T)$.
Define $\delta_{t} = r_t + \gamma \hat{V}_{\bm{\psi}}^{\pi_{\bm{\theta}_{k}}}(\bm{s}_{t+1}) - \hat{V}_{\bm{\psi}}^{\pi_{\bm{\theta}_{k}}}(\bm{s}_t)$, then an estimate of $A^{\pi_{\bm{\theta}_{k}}}$, denoted by $\hat{A}^{\pi_{\bm{\theta}_{k}}}$, can be computed as
\begin{align} \label{eq:adv_func}
	\hat{A}^{\pi_{\bm{\theta}_{k}}}(\bm{s}_t, a_t) = \sum_{l=0}^{T - t - 1} (\gamma \lambda)^l \delta_{t + l}.
\end{align}
The complete procedure of the PPO algorithm is summarized in Algorithm \ref{algo:ppo}.

\begin{algorithm}[!t]
	\SetAlgoLined
	\DontPrintSemicolon
	\KwData{$D, T, K, \epsilon, \gamma, \lambda$}
	\KwResult{$\pi$}
	Randomly initialize $\bm{\theta}_0$ and $\bm{\psi}_0$\;
	\For{$k = 0, \cdots, K-1$}{
		Collect $D$ state transition trajectories by running policy $\pi_{\bm{\theta}_k}$ for $T$ time instants in each trajectory\;
		Update state value function parameter $\bm{\psi}_{k+1}$ by solving \eqref{eq:state_value_func}\;
		Estimate advantage function via \eqref{eq:adv_func}\;
		Update policy parameter $\bm{\theta}_{k+1}$ by solving \eqref{eq:policy_update}\;
	}
	\label{algo:ppo}
	\caption{PPO-based Policy Learning \cite{schulman2017proximal}}
\end{algorithm}

\section{Numerical Simulation} \label{sec:simu}

\begin{figure}[!t]
\centering
\includegraphics[width=3.5in]{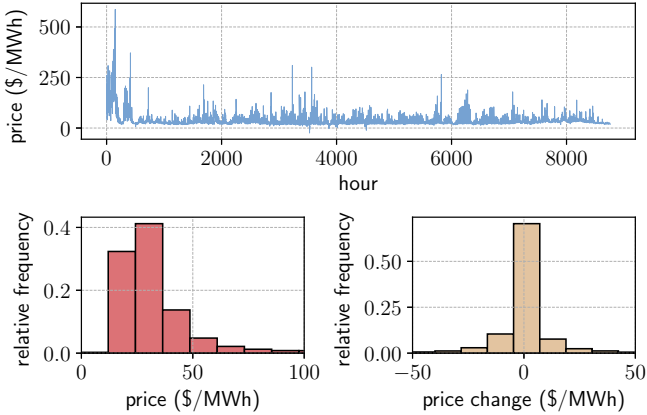}
\caption{Sequence (upper) and histograms of electricity prices (lower left) and price changes (lower right) during 2018 in PJM.}
\label{fig:PJM_price}
\end{figure}

We next demonstrate the effectiveness of the proposed algorithm using actual real-time electricity prices from PJM \cite{PJM}.
Figure \ref{fig:PJM_price} shows the sequence as well as histograms of electricity prices during 2018.
Electricity prices from the first $9$ months and the last $3$ months are used as the training and testing data, respectively.
An EMA filter with $\alpha = 0.7$ and a one-layer RNN with $n = 16$ units are used to extract the hidden state.
The RNN is trained on the training data for $4000$ steps with a learning rate of $0.01$ using the ADAM algorithm \cite{kingma2014adam}.
Both functions $\mu$ and $\hat{V}^\pi_{\bm{\psi}}$ are represented by neural networks with two hidden layers with $128$ and $32$ units each, and rectified linear units as the activation function.
No activation function is used in the output layer.
We perform $K = 200$ updates.
Before each update, $D = 10$ trajectories, each of which has a length $T = 168$ time instants (corresponding to one week) is collected.
Equivalently, the algorithm is trained using data of $2000$ weeks, which is obtained via sampling with replacement.
In each update, \eqref{eq:state_value_func} and \eqref{eq:policy_update} are solved using the ADAM algorithm for $100$ steps with respective learning rates of $1\times 10^{-3}$ and $1\times 10^{-4}$.
Other parameters are set as follows: $\underline{E} = 0$, $\overline{E} = 8$~MWh, $\overline{p}^d = \overline{p}^c = 2$~MW, $\eta^d = \eta^c = 1$, $\tau = 1$~hour, $\beta = 1$~\$/MWh, $\gamma = 0.999$, $\lambda = 0.97$, $\epsilon = 0.2$.

\begin{figure}[!t]
\centering
\includegraphics[width=3.5in]{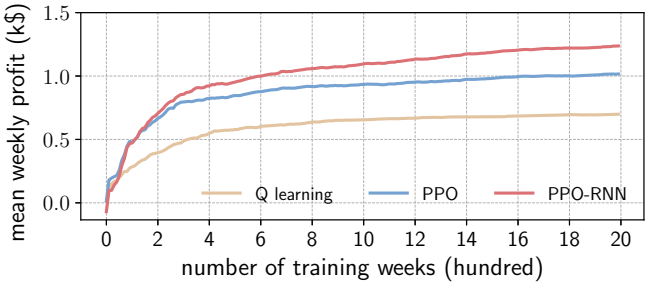}
\caption{Mean weekly profits during training process.}
\label{fig:train_episode_return}
\end{figure}

\begin{figure}[!t]
\centering
\includegraphics[width=3.5in]{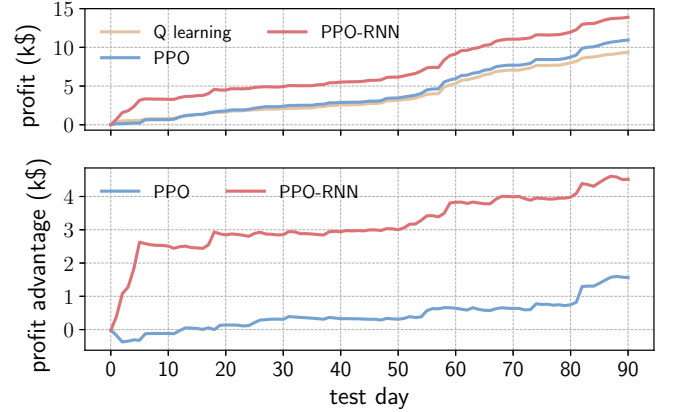}
\caption{Cumulative profits (upper) and cumulative profit advantages of PPO and PPO-RNN over Q learning (lower) during test.}
\label{fig:cum_profit_comp}
\end{figure}

The proposed algorithm is benchmarked against a well-tuned version of the Q learning algorithm proposed in \cite{wang2018energy}, in which the electricity prices and the energy levels are discretized into $100$ and $10$ intervals, respectively.
Figure \ref{fig:train_episode_return} shows the mean weekly profit $\sum_{t=1}^{168} (\rho_t \eta^d - c_t) p^d_t \tau$ (recall that one week corresponds to one trajectory) as the number of training weeks increases, where the proposed algorithm without hidden state extraction is labeled as PPO, and the one with hidden state extraction is labeled as PPO-RNN.
The cumulative profit obtained during testing, and the profit advantages of the proposed algorithm over the Q learning algorithm are presented in Fig. \ref{fig:cum_profit_comp}.
The profits obtained by the Q learning, PPO, and PPO-RNN algorithms during the last $3$ months in 2018 are \$$9377$, \$$10942$, \$$13892$, respectively.
We also evaluate these algorithms under the setup using electricity prices during 2016 and 2017.
The profits obtained by the Q learning, PPO, and PPO-RNN algorithms are respectively \$$6119$, \$$7383$, \$$8750$ during the last $3$ months in 2016, and \$$6371$, \$$7818$, \$$8704$ during the last $3$ months in 2017.
In all cases, the PPO-RNN algorithm obtains approximately $40\%$ more profits than the Q learning algorithm.


\section{Concluding Remarks} \label{sec:con}

In this letter, we proposed a DRL based algorithm for controlling ESSs to arbitrage in real-time electricity markets under price uncertainty.
The proposed algorithm utilizes information extracted from electricity price sequences by an EMA filter and an RNN, and learns an effective stochastic control policy for ESSs.
Numerical simulations using actual electricity prices demonstrated the good performance of the proposed algorithm.

\bibliographystyle{IEEEtran}
\bibliography{ESS_DRL}

\begin{thebibliography}{1}
\providecommand{\url}[1]{#1}
\csname url@samestyle\endcsname
\providecommand{\newblock}{\relax}
\providecommand{\bibinfo}[2]{#2}
\providecommand{\BIBentrySTDinterwordspacing}{\spaceskip=0pt\relax}
\providecommand{\BIBentryALTinterwordstretchfactor}{4}
\providecommand{\BIBentryALTinterwordspacing}{\spaceskip=\fontdimen2\font plus
\BIBentryALTinterwordstretchfactor\fontdimen3\font minus
  \fontdimen4\font\relax}
\providecommand{\BIBforeignlanguage}[2]{{%
\expandafter\ifx\csname l@#1\endcsname\relax
\typeout{** WARNING: IEEEtran.bst: No hyphenation pattern has been}%
\typeout{** loaded for the language `#1'. Using the pattern for}%
\typeout{** the default language instead.}%
\else
\language=\csname l@#1\endcsname
\fi
#2}}
\providecommand{\BIBdecl}{\relax}
\BIBdecl

\bibitem{eyer2010energy}
J.~Eyer and G.~Corey, ``Energy storage for the electricity grid: Benefits and
  market potential assessment guide,'' \emph{Sandia National Laboratories},
  vol.~20, no.~10, p.~5, 2010.

\bibitem{krishnamurthy2018energy}
D.~Krishnamurthy, C.~Uckun, Z.~Zhou, P.~R. Thimmapuram, and A.~Botterud,
  ``Energy storage arbitrage under day-ahead and real-time price uncertainty,''
  \emph{IEEE Trans. Power Syst.}, vol.~33, no.~1, pp. 84--93, 2018.

\bibitem{wang2018energy}
H.~Wang and B.~Zhang, ``Energy storage arbitrage in real-time markets via
  reinforcement learning,'' in \emph{Proc. of IEEE Power \& Energy Society
  General Meeting}, 2018, pp. 1--5.

\bibitem{xu2019deep}
H.~Xu, H.~Sun, D.~Nikovski, S.~Kitamura, K.~Mori, and H.~Hashimoto, ``Deep
  reinforcement learning for joint bidding and pricing of load serving
  entity,'' \emph{IEEE Trans. on Smart Grid}, 2019.

\bibitem{schulman2017proximal}
J.~Schulman, F.~Wolski, P.~Dhariwal, A.~Radford, and O.~Klimov, ``Proximal
  policy optimization algorithms,'' \emph{arXiv preprint arXiv:1707.06347},
  2017.

\bibitem{sutton2018reinforcement}
R.~S. Sutton and A.~G. Barto, \emph{Reinforcement Learning: An
  Introduction}.\hskip 1em plus 0.5em minus 0.4em\relax MIT press, 2018.

\bibitem{goodfellow2016deep}
I.~Goodfellow, Y.~Bengio, A.~Courville, and Y.~Bengio, \emph{Deep
  Learning}.\hskip 1em plus 0.5em minus 0.4em\relax MIT press Cambridge, 2016.

\bibitem{PJM}
``{PJM} hourly {LMP},''
  \url{https://dataminer2.pjm.com/feed/rt\_da\_monthly\_lmps}.

\bibitem{kingma2014adam}
D.~P. Kingma and J.~Ba, ``Adam: A method for stochastic optimization,''
  \emph{arXiv preprint arXiv:1412.6980}, 2014.

\end{thebibliography}

\end{document}